\documentclass[letterpaper, 10 pt, conference]{ieeeconf}
\IEEEoverridecommandlockouts       
\usepackage{graphics}
\usepackage{epsfig}
\usepackage{mathptmx}
\usepackage{times}
\usepackage{amsmath}
\usepackage{amssymb}
\usepackage{multirow}
\usepackage{hhline}
\usepackage{booktabs}
\usepackage{float}
\usepackage{array}
\usepackage{makecell}
\usepackage{lipsum}
\usepackage{caption}
\usepackage{subcaption}
\usepackage{subcaption}\usepackage[font=footnotesize]{caption}
\usepackage{hyperref}
\usepackage{soul}
\hypersetup{
    colorlinks=true,
    linkcolor=blue,
    filecolor=magenta,      
    urlcolor=blue,
}
\DeclareMathOperator*{\argmax}{arg\,max}
\DeclareMathOperator*{\argmin}{arg\,min}

\title{\LARGE \bf
Active Extrinsic Contact Sensing:\\Application to General Peg-in-Hole Insertion}
\vspace{-3ex}
\author{Sangwoon Kim, Alberto Rodriguez

\thanks{This research is supported by the HKSAR Innovation and Technology Fund (ITF) ITS-104-19F, and by the Mitsubishi Electric Research Laboratory (MERL).}% <-this % stops a space
\thanks{Sangwoon Kim and Alberto Rodriguez are with the Department of Mechanical Engineering, Massachusetts Institute of Technology, {\tt\small <sangwoon, albertor>@mit.edu}}
}

\begin{document}

\maketitle
\thispagestyle{empty}
\pagestyle{empty}

%%%%%%%%%%%%%%%%%%%%%%%%%%%%%%%%%%%%%%%%%%%%%%%%%%%%%%%%%%%%%%%%%%%%%%%%%%%%%%%%
\begin{abstract}

    We propose a method that actively estimates contact location between a grasped rigid object and its environment and uses this as input to a peg-in-hole insertion policy. An estimation model and an active tactile feedback controller work collaboratively to estimate the external contacts accurately. The controller helps the estimation model get a better estimate by regulating a consistent contact mode. The better estimation makes it easier for the controller to regulate the contact. We then train an object-agnostic insertion policy that learns to use the series of contact estimates to guide the insertion of an unseen peg into a hole. In contrast with previous works that learn a policy directly from tactile signals, since this policy is in contact configuration space, it can be learned directly in simulation. Lastly, we demonstrate and evaluate the active extrinsic contact line estimation and the trained insertion policy together in a real experiment. We show that the proposed method inserts various-shaped test objects with higher success rates and fewer insertion attempts than previous work with end-to-end approaches. See supplementary video and results at \href{https://sites.google.com/view/active-extrinsic-contact}{https://sites.google.com/view/active-extrinsic-contact}.

\end{abstract}

%%%%%%%%%%%%%%%%%%%%%%%%%%%%%%%%%%%%%%%%%%%%%%%%%%%%%%%%%%%%%%%%%%%%%%%%%%%%%%%%

\section{Introduction}

Sensing and utilizing tactile feedback between fingers and grasped objects is key to dexterous manipulation skills \cite{Howe}. Tactile sensing can be used as a feedback signal to regulate desired contact configurations \cite{Rodriguez}. One major problem in tactile feedback control is localizing and controlling an external contact between the grasped object and its environment. For example, consider pivoting an unknown object resting on a surface while avoiding slipping at the contact between the object and the surface. It requires an ability to regulate the external contact using indirect observations, possibly through tactile sensing.

In the peg-in-hole insertion task, the external contact matters. During the insertion attempt, a misalignment between the peg and the hole leads to unintended contact. The contact triggers a tactile signal, which can be used to localize the contact or plan the next insertion attempt. Key challenges in this task are as follows:
\begin{itemize}
\item The tactile signal is a partial observation of the contact state; many different contact configurations can cause the same tactile signal \cite{Newman}.
\item The frictional contact mechanics that govern the alignment and insertion dynamics are sensitive to switching contact modes.
%\item For learning-based methods, it should have a reasonably high data efficiency.
\end{itemize}

This work tackles both the extrinsic contact localization problem and the peg-in-hole insertion task in a combined framework. For the extrinsic contact localization, we use a factor graph solved with incremental smoothing and mapping (iSAM) \cite{Kaess2008} to fuse the information of robot proprioception and tactile measurements. The factor graph works collaboratively with an active tactile feedback controller that attempts to pivot the object about a regulated external contact to generate sufficient observations to estimate the contact line between an unknown object and an unknown environment. The contact line estimation is then used as an input to the insertion policy, as opposed to the end-to-end approach where the policy takes directly as input the raw tactile feedback. Since the input to the policy is a low dimensional representation (a contact line), the policy training can be done in a simple 2D geometric simulation, so there is no need to collect training data in a real experiment. Lastly, we demonstrate and evaluate the extrinsic contact line estimation and the trained insertion policy in a set of real experiments.

\begin{figure}[t]
	\centering
	\includegraphics[width=0.8\linewidth]{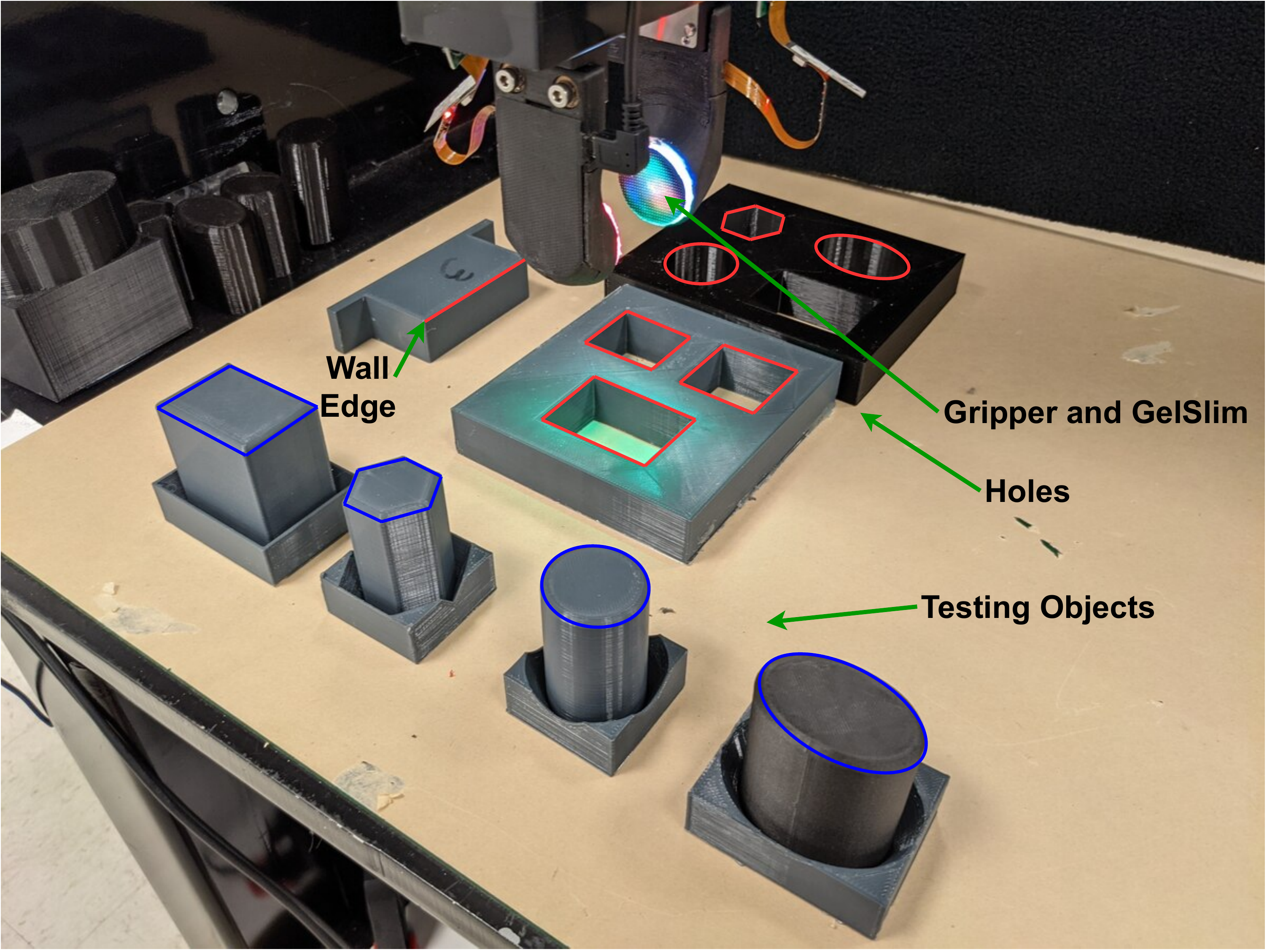}\vspace{0pt}
	\caption{Experimental setup: Gripper with GelSlim \cite{Taylor}, different objects (blue), and different environments (red)}
	\label{fig:setup}
	\vspace{-15pt}
\end{figure}

\begin{figure*}[t]
	\centering
	\includegraphics[width=.85\linewidth]{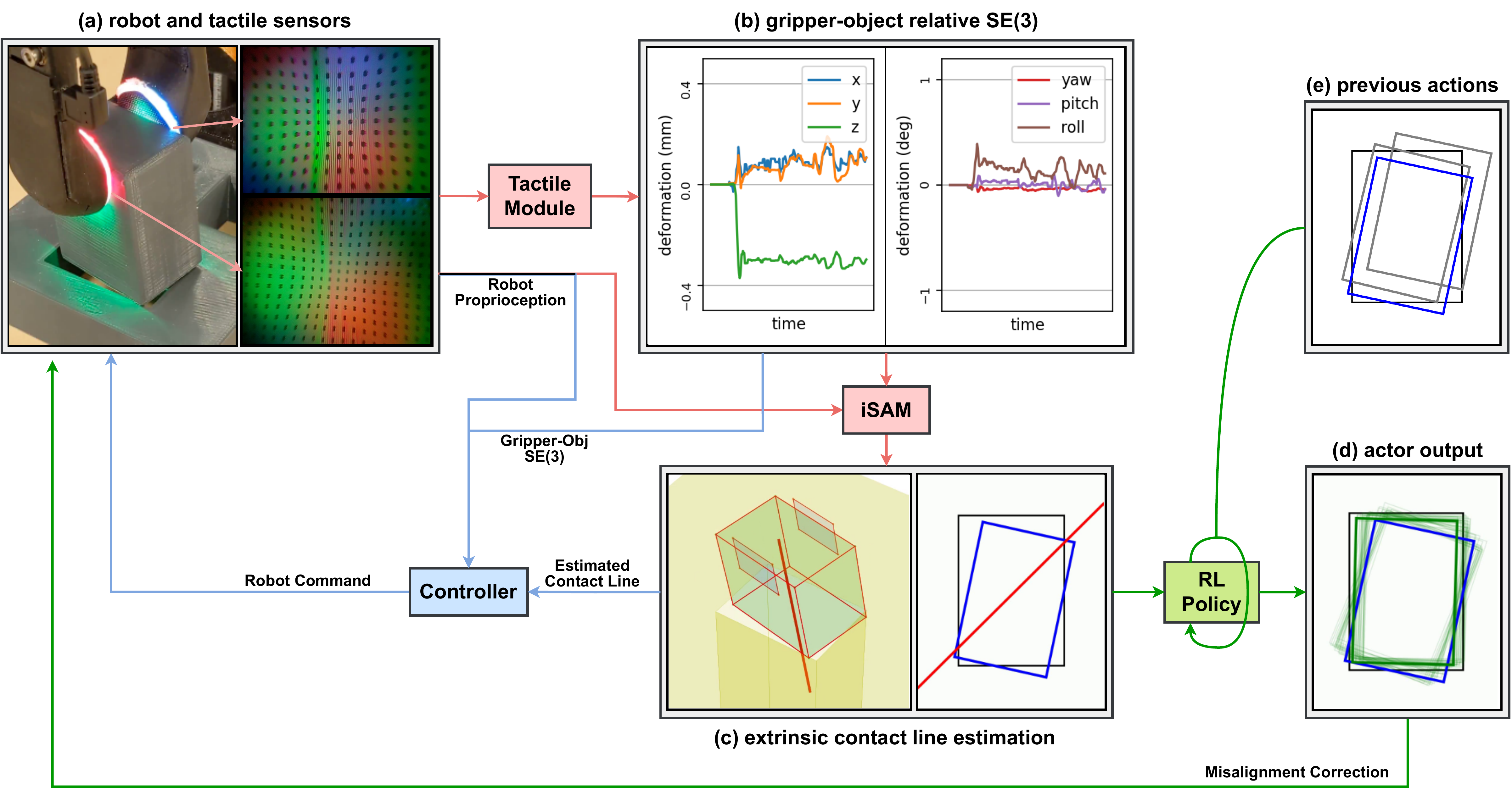}\vspace{-0pt}
	\caption{Approach Overview. Estimation module (red) and active tactile feedback controller (blue) run collaboratively to estimate extrinsic contact line. RL policy (green) takes the estimated extrinsic contact line as input and computes the next action. (a) Insertion attempt and tactile images captured by GelSlim fingers. (b) Gripper-Object relative displacement computed by a learned tactile module. (c) 3D and top view of the extrinsic contact line estimation. The bold red line is the current estimate. (d) RL Actor output. The blue rectangle is the current pose and the green rectangles are the candidate poses for the next insertion attempt. (e) History of previous attempts that feed into the recurrent RL architecture.}
	\label{fig:overview}
	\vspace{-15pt}
\end{figure*}

\section{Related Work}

\subsection{Tactile Sensing and Feedback}

Prior work showed that tactile measurements could be used for state estimation. Bicchi, et al. \cite{Bicchi} used force measurements to estimate a contact location when the force is exerted on the robot with known geometry. Yu and Rodriguez \cite{Yu2018ICRA} combined force and visual sensing to estimate the pose of a known planar object in planar manipulation. They also used a similar framework to estimate the SE(3) pose of a known object in touch with an external environment \cite{Yu2018IROS}.

Vision-based tactile sensors like GelSight \cite{Yuan} and GelSlim \cite{Donlon,Taylor} enabled accurate estimation of contact states. These sensors capture the physical interaction between the robot fingers and the grasped object as high-resolution images. Bauza, et al. \cite{Bauza2019, Bauza2020} used tactile images to map and localize the relative pose of a known grasped object. Ma, et al. \cite{Ma2019} conducted inverse finite element method (iFEM) on tactile images to reconstruct a dense tactile force distribution. Sodhi, et al. \cite{Sodhi} developed a factor graph model that finds a maximum a posteriori (MAP) estimate of an end-effector and an object pose in planar manipulation with a vision-based tactile sensor \cite{Lambeta}. While the work above focused on estimating the states of an object or an interface in direct touch with a tactile sensor, Ma, et al. \cite{Ma2021} focused on localizing extrinsic contacts. They accumulated a sequence of tactile images and used least-squares fitting \cite{Arun} to estimate the location of the extrinsic contact. However, all these estimation methods are passive, so they are limited to using the information they are presented with.

The tactile sensing can be used as a feedback signal to regulate the desired contact configuration. Dong, et al. \cite{Dong2019ICRA} introduced a module that monitors incipient slip on the tactile sensor and uses it to maintain a stable grasp. Hogan, et al. \cite{Hogan} developed a closed-loop tactile controller for dexterous manipulation primitives. She, et al. \cite{She} showed tactile sensing for manipulating a cable. They learned a linear model for the cable sliding dynamics and implemented a linear quadratic regulator (LQR) to keep the cable near the sensor center while sliding through the cable.

\subsection{Peg-in-Hole Insertion}

Early studies on peg-in-hole insertion relied on passive compliance of the gripper \cite{Whitney}. Other works assume a known object-hole model \cite{Came, Bruyninckx}. These methods are object-specific and difficult to apply to unknown objects. More recently, model-free, end-to-end, learning-based approaches have been proposed \cite{Inoue, Lee, Dong2019IROS, Dong2021}. \cite{Inoue} used deep reinforcement learning (RL) for a high precision peg-in-hole task. They used a force/torque (F/T) sensor measurement and a robot position as input to a discrete action policy. \cite{Lee} fused force and vision to address the peg-in-hole insertion task. \cite{Dong2019IROS} used supervised learning to map a tactile image sequence to a misalignment between peg and hole. However, since the tactile image does not fully observe the state, their approach showed sub-optimal performance. The same group of researchers used end-to-end RL to overcome this limitation and showed  a performance improvement \cite{Dong2021}. Simulating raw images is difficult in these end-to-end approaches, so the training data was collected in real experiments.

\begin{figure}[t]
	\centering
	\includegraphics[width=0.9\linewidth]{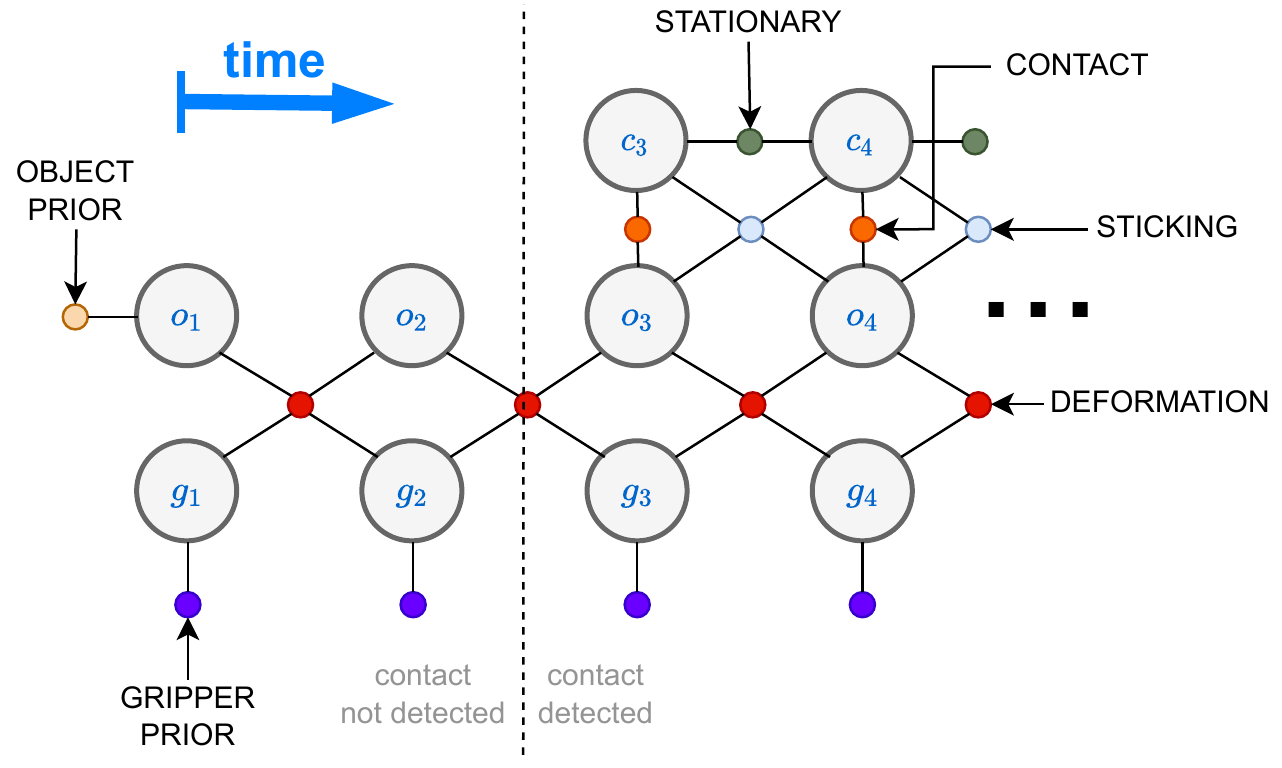}\vspace{0pt}
	\caption{Estimation factor graph}
	\label{fig:graph}
	\vspace{-5pt}
\end{figure}

\begin{figure}[t]
	\centering
	\includegraphics[width=.7\linewidth]{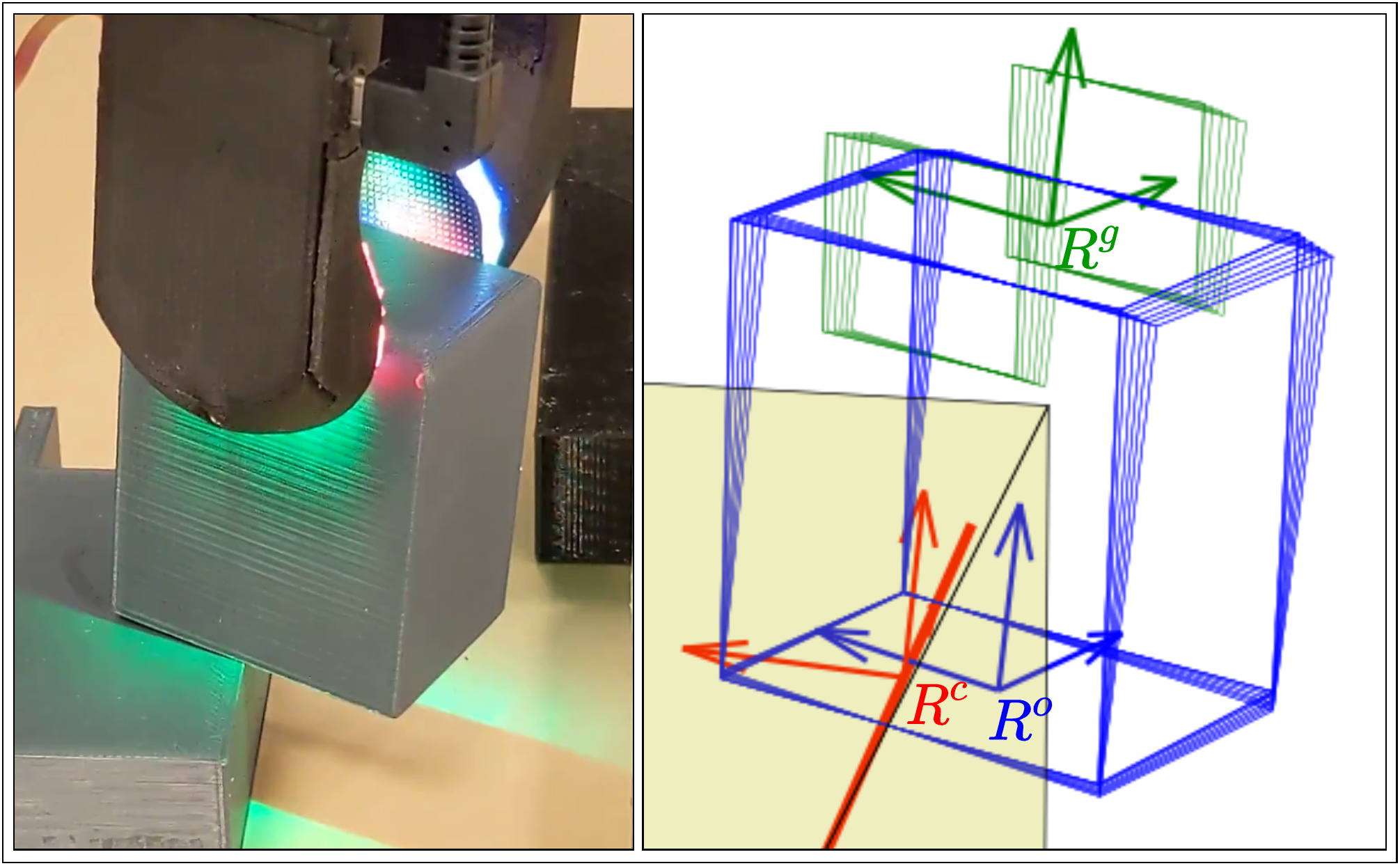}\vspace{0pt}
	\caption{Left: An object in contact with an edge. Right: Gripper (green) and object (blue) trajectories and an estimated contact line (red).}
	\label{fig:cline}
	\vspace{-15pt}
\end{figure}

\section{Methodology}

The task we solve is a typical peg-in-hole problem. We make several assumptions to implement our framework:
\begin{itemize}
    \item Object bottom surfaces and hole top surfaces are flat.
    \item The misalignment between object and hole is an SE(2) displacement in the plane of contact.
    \item Objects and holes are un-chamfered.
\end{itemize}

The approach comprises two parts: the active extrinsic contact sensing and the insertion policy. In the active extrinsic contact sensing, a factor graph estimator works collaboratively with an active tactile feedback controller to estimate the extrinsic contact line. The controller helps the factor graph improve the estimation by regulating a consistent extrinsic contact mode. The improved estimation from the factor graph helps the controller better regulate the extrinsic contact. Then, the insertion policy learned in simulation with reinforcement learning (RL) takes the estimated contact line as input and computes the next action for the insertion.

\subsection{Tactile Module and Factor Graph} \label{subsection:iSAM}

We use GelSlim 3.0 \cite{Taylor}, a vision-based tactile sensor, to capture the deformation image on the robot finger during the insertion (Fig.\ref{fig:overview}a). The image is passed to a convolutional neural network (CNN) architecture, the tactile module, trained with supervised learning to estimate the relative SE(3) displacement between the gripper and the object due to the compliance of the finger (Fig.\ref{fig:overview}b). The gripper-object relative displacement and the robot proprioception data are then used in the factor graph to infer the extrinsic contact line (Fig.\ref{fig:overview}c).

Fig.\ref{fig:graph} shows the factor graph. Each color of small circles represents different types of factors. $g_i$, $o_i$, and $c_i$ are the SE(3) gripper pose, object pose, and contact line as shown in Fig.\ref{fig:cline} ; $xy$ surface of $o_i$ represents the bottom surface of the object and $x$-axis of $c_i$ represents the estimated contact line. iSAM solver computes a maximum a posteriori (MAP) estimate for trajectories of the gripper pose $G$, the object pose $O$, and the contact line $C$, given the gripper pose measurements based on robot proprioception $R$, the gripper-object relative displacement $D$, and the initial object pose prior $b_o$:
\vspace{-1.43pt}
\begin{align}
    G^*, O^*, C^* &= \argmax_{G, O, C} P(G, O, C, R, D, b_o)% \nonumber\\
    %=& \argmin_{G, O, C} -\log{P(G, O, C, R, D, b_o)}.
\end{align}\vspace{-1.43pt}
Assuming Gaussian noise, solving for the MAP estimate becomes a nonlinear least-squares problem:
\vspace{-1.43pt}
\begin{align}
    G^*,& O^*, C^* = \argmin_{G, O, C}  \{ \sum_{t=1}^{T}\{||F_{gp}(g_t,r_t)||_{\sum_{gp}}^2 \nonumber\\
    & + ||F_{def}(g_{t-1}, g_t ,o_{t-1}, o_t, d_t)||_{\sum_{def}}^2 + ||F_{ct}(o_t ,c_t)||_{\sum_{ct}}^2 \nonumber\\
    & + ||F_{tic}(o_{t-1}, o_t ,c_{t-1}, c_t)||_{\sum_{tic}}^2 + ||F_{stat}(c_{t-1}, c_t)||_{\sum_{stat}}^2\} \nonumber\\
    & + ||F_{op}(o_1,b_o)||_{\sum_{op}}^2 \},
\end{align}\vspace{-1.43pt}
where $F(\cdot)$ are cost functions for each factor.
%Each term represents the gripper prior factor $F_{gp}(\cdot)$, GelSlim deformation factor $F_{def}(\cdot)$, contact factor $F_{ct}(\cdot)$, sticking factor $F_{tic}(\cdot)$, stationary factor $F_{stat}(\cdot)$, and object prior factor $F_{op}(\cdot)$.
$||\cdot||_{\sum}$ is the Mahalanobis distance with covariance $\sum$. iSAM enables to add new measurements incrementally and update the estimation in real-time rather than solving it from scratch at every step.

\textbf{Gripper prior ($F_{gp}$) and Object prior ($F_{op}$)}: We use unary factors to model the uncertainty of the gripper pose and initial object pose:
\vspace{-1.43pt}
\begin{align}
    ||F_{gp}(g_t,r_t)||_{\sum_{gp}}^2 :=& ||g_t^{-1} r_t||_{\sum_{gp}}^2\\
    ||F_{op}(o_1,b_o)||_{\sum_{op}}^2 :=& ||o_1^{-1} b_o||_{\sum_{op}}^2,
\end{align}\vspace{-1.43pt}
where $r_t$ is the measured gripper pose based on robot proprioception and $b_o$ is the prior knowledge about the initial object pose.

\textbf{GelSlim deformation factor ($F_{def}$)}: Relative SE(3) displacement between gripper and object computed by the tactile module is incorporated as the GelSlim deformation factor:
\vspace{-1.43pt}
\begin{align}
    ||F_{def}(g_{t-1},& g_t, o_{t-1}, o_t, d_t)||_{\sum_{def}}^2 \nonumber\\
    & := ||((g_{t-1}^{-1}o_{t-1})^{-1}(g_t^{-1}o_t))^{-1} d_t||_{\sum_{def}}^2,
\end{align}\vspace{-1.43pt}
where $d_t$ is the change in the relative displacement from $t-1$ to $t$.

\textbf{Contact factor ($F_{ct}$)}: To constrain the estimated contact line to lie on the object bottom surface, we use a binary factor:
\vspace{-1.43pt}
\begin{equation}
    ||F_{ct}(o_t, c_t)||_{\sum_{ct}}^2 := ||(o_t^{-1} c_t)_{z^c, R_{x^c}, R_{y^c}}||_{\sum_{ct}}^2,
\end{equation}\vspace{-1.43pt}
where the subscript $z^c, R_{x^c}, R_{y^c}$ indicates that we only constrain the components that move out of the object's bottom surface.

\begin{figure}[t]
	\centering
	\includegraphics[width=0.45\linewidth]{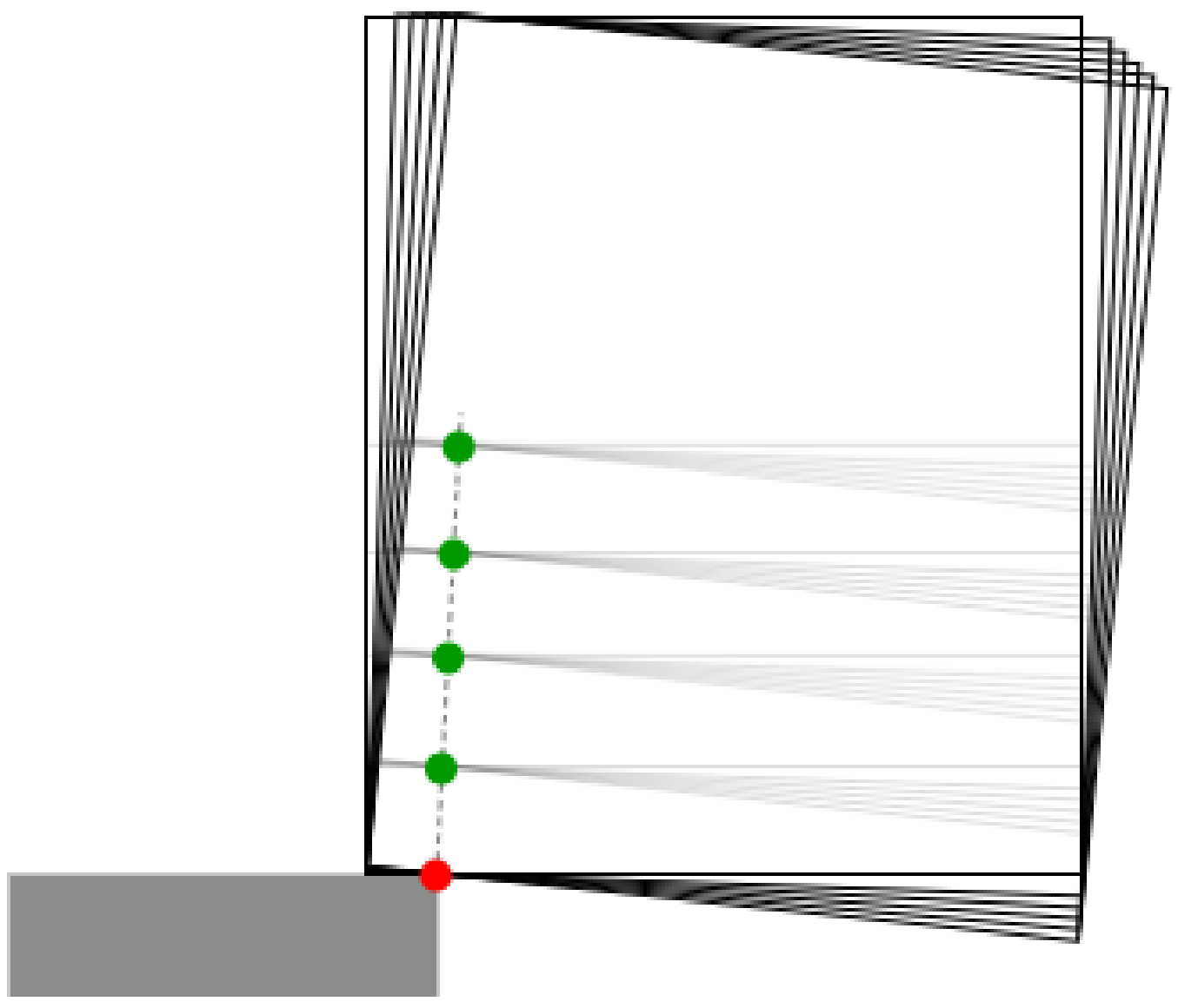}
	\caption{An object in touch with an environment. The sticking factor enables the factor graph to distinguish red from green points.}
	\vspace{-15pt}
    \label{fig:stick}
\end{figure}

\textbf{Sticking factor ($F_{tic}$)}: Fig.\ref{fig:stick} shows an example of an object touching an environment and tilting at a small angle. In such a case, it is difficult to infer the actual contact point with only geometric constraints. All the green points in the figure will seem to satisfy the geometric constraints for a certain object height. Therefore, while we can get a reasonable estimate in the horizontal direction, the estimated height of the contact will be uncertain. However, if we use a controller to regulate the external contact to be sticking, only the red dot in Fig.\ref{fig:stick} is now compatible with the constraints. We incorporate it in the factor graph and get a better estimate in the vertical direction. We use a binary factor to impose the sticking constraint:
\vspace{-1.43pt}
\begin{equation}
    ||F_{tic}(o_{t-1}, o_t ,c_{t-1}, c_t)||_{\sum_{tic}}^2 := ||((o_{t-1}^{-1}c_{t-1})^{-1}(o_t^{-1}c_t))_{y^c}||_{\sum_{tic}}^2,
\end{equation}\vspace{-1.43pt}
where the subscript $y^c$ is the direction on the object's bottom surface perpendicular to the contact line.

\textbf{Stationary factor ($F_{stat}$)}: We assume contact configuration does not change during the motion and the contact line is stationary. This is incorporated by using a binary factor between consecutive timesteps:
\vspace{-1.43pt}
\begin{equation}
    ||F_{stat}(c_{t-1}, c_t)||_{\sum_{stat}}^2 := ||(c_{t-1}^{-1}c_t)_{y^c,z^c,R_{y^c},R_{z^c}}||_{\sum_{stat}}^2
\end{equation}\vspace{-1.43pt}

\begin{figure}[t]
	\centering
	\includegraphics[width=\linewidth]{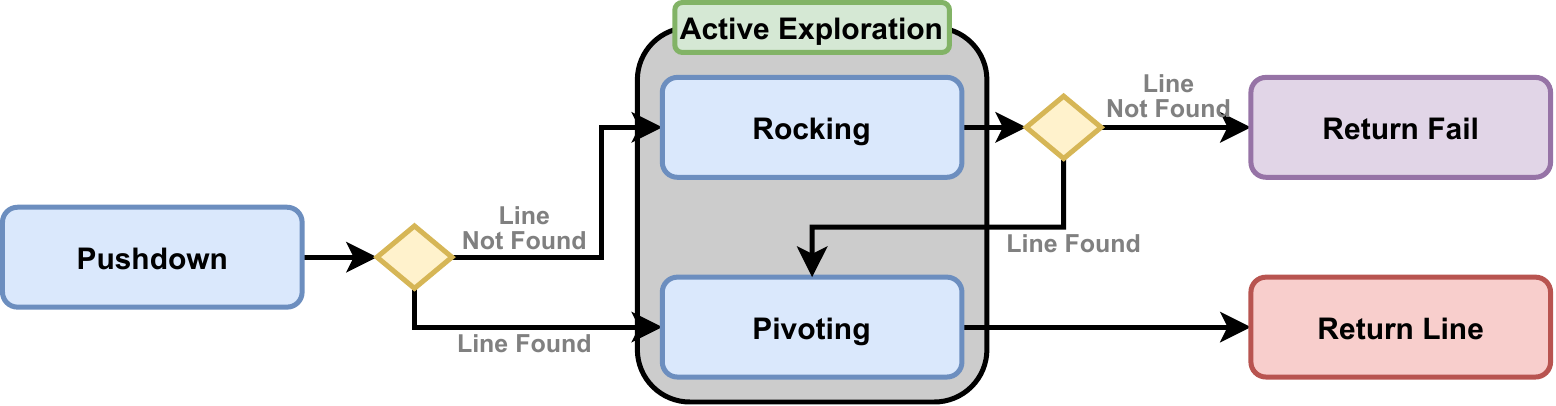}\vspace{0pt}
	\caption{Active tactile feedback controller}
	\vspace{-5pt}
	\label{fig:control}
\end{figure}

\vspace{-12pt}\subsection{Active tactile feedback Controller}

Fig.\ref{fig:control} shows the flowchart of the active tactile feedback controller. It comprises a push-down phase where the robot moves down until detecting contact, and an active exploration phase where the robot tries to rock and pivot the object about the external contact without sliding. During the push-down phase, it executes a proportional control on the gripper-object relative pose estimated by the tactile module:
\vspace{-1.43pt}
\begin{align}
    v_z^g &= -K_{p,z}(\Delta z_d^g-\Delta z^g) ~~~~~~ \text{(vertical)} \\
    \omega_x^g &= -K_{p,\phi}(\Delta \phi_d^g-\Delta \phi^g) ~~~~~ \text{(roll)} \\
    \omega_y^g &= -K_{p,\theta}(\Delta \theta_d^g - \Delta \theta^g) ~~~~~ \text{(pitch)},
\end{align}\vspace{-1.43pt}
where $[\omega_x^g, \omega_y^g, \omega_z^g, v_x^g, v_y^g, v_z^g]^T = [\omega^g, \mathbf{v}^g]^T$ is the body twist of the gripper frame and the $\Delta$'s are the relative displacement from when the object was not in contact with the environment. The subscript $d$ is for the desired relative displacement. $\Delta z_d^g$ is set to a non-zero value to ensure the object contacts the environment with sufficient normal force. $\Delta \phi_d^g$ and $\Delta \theta_d^g$ are set to zero.

If the object does not tilt enough so the factor graph fails to estimate a contact line with enough confidence, it enters the rocking phase. In the rocking phase, the robot follows a cone-like trajectory by setting the desired relative displacement as below:
\vspace{-1.43pt}
\begin{equation}
    (\Delta \phi_d^g, \Delta \theta_d^g)(t) = \Delta \phi_0^g (\cos{\omega t}, \sin{\omega t}) \nonumber
\end{equation}\vspace{-1.43pt}
If the factor graph fails to find a contact line even after the rocking, the controller stops and the estimator returns a failure signal.

If the factor graph succeeds in estimating an extrinsic contact line either in push-down or rocking, the controller enters the pivoting phase to help the factor graph to get more accurate estimate. The controller tries to pivot the object around the extrinsic contact line while avoiding slipping and maintaining a constant tactile deformation. It is assumed that the tactile deformation will remain constant during the pivoting if the object pivots without slipping and maintains a constant contact force. Proportional control is used to maintain the deformation as constant as possible:
\vspace{-1.43pt}
\begin{align}
    \omega^g &= \pm \omega_0 \hat{x}^{gc} - K_{p,\beta} (\Delta \beta_d^g - \Delta \beta^g) \hat{y}^{gc}, \label{eq:rock1} \\
    \mathbf{v}^g &= \mathbf{p}^{gc} \times \omega^g - K_{p,\alpha} (\Delta \alpha_d^g - \Delta \alpha^g) \hat{y}^{gc} \nonumber\\
    &\qquad\qquad- K_{p,z}(\Delta z_d^g-\Delta z^g) \hat{z}^g, \label{eq:rock2}
\end{align}\vspace{-1.43pt}
where $\hat{x}^{gc}$ and $\hat{y}^{gc}$ are the $\hat{x}^{c}$ and $\hat{x}^{c}$ in the gripper coordinate. $\Delta \alpha^g$ and $\Delta \beta^g$ are the components of the relative rotation change in the $\hat{x}^{c}$ and $\hat{y}^{c}$ direction. $\omega_0$ is the rotational speed of the pivoting. $\mathbf{p}^{gc}$ is the origin of the contact line from the gripper coordinate. The desired relative displacement ($\Delta_d$) is 
%set to the relative pose change right before the pivoting starts then it is 
updated to a current relative displacement whenever the pivoting direction changes. The first terms of Eq.\ref{eq:rock1} and Eq.\ref{eq:rock2} rotate the gripper around the current contact line estimate and alter direction ($+\omega_0$ and $-\omega_0$) as it pivots the object back and forth. The second term of Eq.\ref{eq:rock1} and the third term of Eq.\ref{eq:rock2} ensure the object is being pushed down while maintaining the same line contact. The second term of Eq.\ref{eq:rock2} translates the gripper to the direction it reduces slipping.

A key idea in the above method is the synergistic interaction between the controller and the estimator. A better contact line estimation helps the controller pivot the object with less slipping. On the other hand, better pivoting around a consistent axis allows the estimator to get a more accurate contact line estimate. %The interaction acts as a stable attractor that enables to find an accurate contact line estimate with a simple proportional controller.

\subsection{RL Policy}

The RL policy takes the estimated extrinsic contact line from the estimation graph as the input and computes the SE(2) pose correction for the next insertion attempt (Fig.\ref{fig:overview}d). Since the input to the policy is a low dimensional representation (a single contact line), it is computationally trivial to simulate the policy; given two random shape polygons, each representing the object and the hole, and an SE(2) misalignment between the two, it is easy to find a contact line that the object can pivot around as can be seen in Fig.\ref{fig:overview}c(right). Further details are discussed in Section \ref{section:details}.

\begin{figure}[t]
	\centering
	\includegraphics[width=\linewidth]{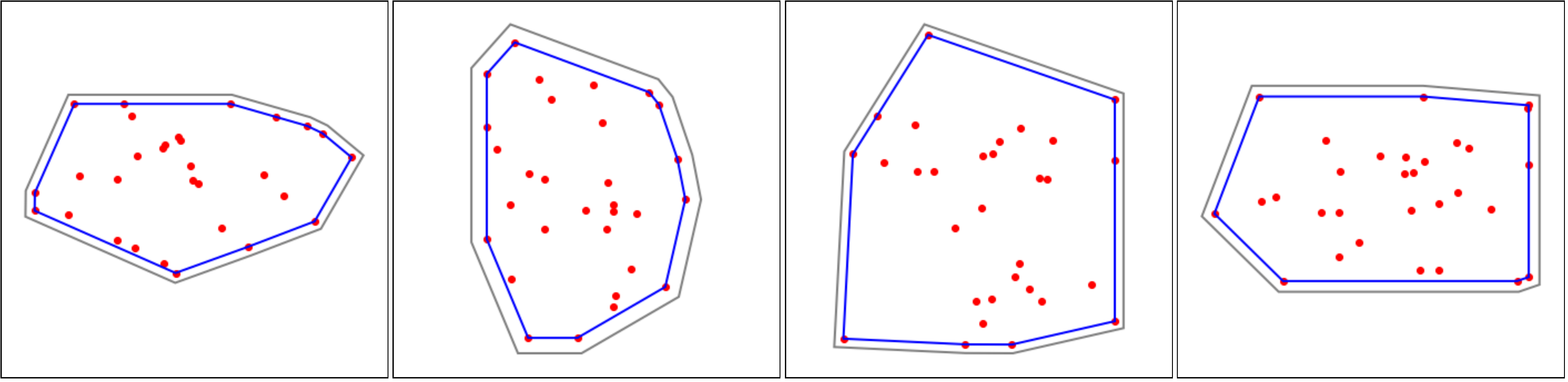}
	\caption{Random polygonal object-holes}
	\vspace{-15pt}
    \label{fig:scatter}
\end{figure}

\section{Experimental Details} \label{section:details}

\subsection{Experimental Setup}

Fig.\ref{fig:setup} shows the experimental setup. We use a 6-DoF ABB 120 robot arm and a WSG-50 parallel jaw gripper. GelSlim 3.0 \cite{Taylor} sensors are mounted on each side of the jaw gripper. Four types of 3-D printed testing objects are used on seven types of holes and one single wall environment. The average width of the objects is 35mm and the clearance between the objects and the holes is 2.25mm. The top view of each object-hole pair is shown in Table \ref{table:result2}. We use Pytorch \cite{Pytorch} for training the tactile module and the RL policy, and iSAM2 solver \cite{Kaess2012} implemented in the GTSAM library \cite{Dellaert2012} for the factor graph optimization.

\subsection{Tactile Module Training}

The tactile module is trained with supervised learning to estimate the gripper-object relative displacement. It takes a pair of reference tactile images from each side of the sensors at one time and a pair of query tactile images at another time then computes the relative displacement change between the two times. We fix the four objects then grasp them with random pose and force. Then we randomly wiggle the gripper with Ornstein-Uhlenbeck process \cite{Uhlenbeck} to collect a training sequence of tactile images. The wiggling motion is clipped with a max limit of (0.25mm, 0.5mm, 0.5mm, 1.2$^{\circ}$, 0.6$^{\circ}$ 0.2$^{\circ}$) in the ($x,y,z$, roll, pitch, yaw) direction. We regrasp the object after every 10 seconds. 100 sequences for each object are collected and combined in one large training set. The trained module estimates the relative displacement in the ($y,z$, roll, pitch, yaw) direction with reasonable accuracy: RMSE of (0.07mm, 0.06mm, 0.2$^{\circ}$, 0.1$^{\circ}$, 0.05$^{\circ}$). However, it showed less accuracy in the $x$ direction, the direction perpendicular to the sensor surface: RMSE of 0.12mm. This is expected since it is the direction where the finger gel skin affords less deformation.

\subsection{Active Extrinsic Contact Sensing Experiment}\label{subsection:wall}

Before testing the entire framework in the insertion task, we decouple only the active extrinsic contact sensing part and test the estimation accuracy on the single wall environment. For every grasp, we randomize the grasping height and force. We also vary the horizontal translation and horizontal rotation misalignment between the object and the wall. We say the misalignment is zero when the wall's edge and the object's $x$-axis match when seen from the top view. Translational error is uniformly sampled from $-12\sim12$ mm and rotational error is sampled from $-90^{\circ}\sim90^{\circ}$.

We compare the performance with some ablation models as seen in Table \ref{table:result1}. Note that `w/o Deformation Factor' means that we do not incorporate deformation information into the factor graph but we still use it as the feedback signal. `w/o Control' means that there is no feedback so the gripper pushes down with no tilting and the contact line estimation will only rely on the passive compliance of the GelSlim, as in the case of the earlier work that studied extrinsic contact sensing \cite{Ma2021}.

\begin{table*}[h]
\caption{Accuracy and average error of the active extrinsic contact sensing}
\begin{center}
\vspace{-12pt}
\begin{tabular}{|c||c|c||c|c|c|}
\bottomrule
& \multicolumn{2}{c||}{Accuracy (\%)} & \multicolumn{3}{c|}{Average Error}\\
\cline{2-6}
& Easy & Difficult & \makecell{Horizontal Translation\\(mm)} & \makecell{Horizontal Rotation\\(deg)} & \makecell{Vertical Translation\\(mm)}\\
\hhline{|=|=|=|=|=|=|}
\makecell{\vspace{-7pt}\\\textbf{Active iSAM}\\\vspace{-7pt}} & 95 & 76 & 1.80 & 5.40 & 3.31\\
\hline
\makecell{\vspace{-7pt}\\w/o Sticking Factor\\\vspace{-7pt}} & 91 & 79 & 2.13 & 5.00 & 11.2\\
\hline
\makecell{\vspace{-7pt}\\w/o Deformation Factor\\\vspace{-7pt}} & 92 & 74 & 1.74 & 5.81 & 3.08\\
\hline
\makecell{\vspace{-9pt}\\w/o Deformation Factor\\and w/o Pivoting Motion\\\vspace{-9pt}} & 79 & 55 & 2.58 & 6.96 & 5.50\\
\hline
\makecell{\vspace{-7pt}\\w/o Pivoting Motion\\\vspace{-7pt}} & 90 & 75 & 2.97 & 6.57 & 6.50\\
\hline
\makecell{\vspace{-9pt}\\w/o Active Motion\\(rocking \& pivoting)\\\vspace{-9pt}} & 88 & 34 & 2.37 & 6.89 & 5.36 \\
\hline
\makecell{\vspace{-9pt}\\w/o Control\\(straight push-down)\\\vspace{-9pt}} & 28 & 14 & 3.70& 13.1 & 5.73\\
\toprule
\end{tabular}
\end{center}
\vspace{-10pt}
\label{table:result1}
\end{table*}

\begin{table*}[h]
\caption{Success Rate and average number of insertion attempts (in parentheses) for various object-hole pairs.}
\begin{center}
\vspace{-12pt}
\begin{tabular}{|c||c|c|c|c|c|c|c|}
\bottomrule
\makecell{Object (blue) \& \\  Hole (black) Shape} &
\begin{minipage}{10mm}
\centering
      \includegraphics[height=10mm]{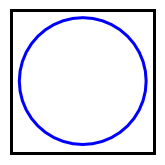}
\end{minipage} &
\begin{minipage}{10mm}
\centering
      \includegraphics[height=10mm]{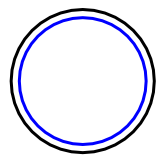}
\end{minipage} &
\begin{minipage}{10mm}
\centering
      \includegraphics[height=10mm]{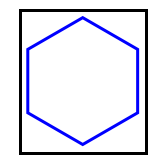}
\end{minipage} &
\begin{minipage}{10mm}
\centering
      \includegraphics[height=10mm]{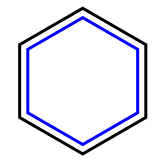}
\end{minipage} &
\begin{minipage}{13mm}
\centering
      \includegraphics[height=10mm]{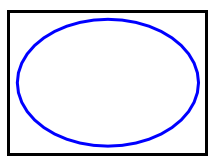}
\end{minipage} &
\begin{minipage}{13mm}
\centering
      \includegraphics[height=10mm]{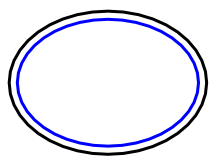}
\end{minipage} &
\begin{minipage}{13mm}
\centering
      \includegraphics[height=10mm]{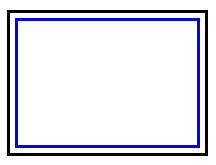}
\end{minipage}\\
\hhline{|=|=|=|=|=|=|=|=|}
\makecell{\textbf{Active} \\ \textbf{iSAM-RL}} & 100\% (1.94) & 97\% (1.98) & 99\% (2.40) & 100\% (2.94) & 95\% (2.61) & 95\% (3.04) & 97\% (4.13)\\
\specialrule{.09em}{0em}{0em}
\makecell{w/o Sticking \\ Factor} & 100\% (1.92) & 100\% (2.02) & 100\% (2.82) & 100\% (2.70) & 96\% (3.15) & 94\% (2.47) & 94\% (3.64)\\
\specialrule{.09em}{0em}{0em}
\makecell{w/o Deformation \\ Factor} & 100\% (2.50) & 94\% (2.32) & 98\% (2.75) & 96\% (3.25) & 92\% (2.83) & 96\% (2.77) & 90\% (4.76)\\
\specialrule{.09em}{0em}{0em}
\makecell{w/o Active \\ Motion} & 96\% (2.15) & 94\% (2.00) & 94\% (2.70) & 94\% (3.51) & 90\% (3.02) & 90\% (2.73) & 86\% (4.77)\\
\specialrule{.09em}{0em}{0em}
\makecell{w/o Control \\ (straight push-down)} & 46\% (4.17) & 46\% (5.87) & 54\% (5.52) & 34\% (4.82) & 58\% (4.62) & 54\% (4.19) & 30\% (5.60)\\
\specialrule{.09em}{0em}{0em}
\makecell{RL-end2end \\ (Dong \cite{Dong2021})} & 97\% (2.96) & - & 97\% (3.83) & - & 98\% (2.34) & - & 90\% (5.42)\\
\specialrule{.09em}{0em}{0em}
\makecell{SL-end2end \\ (Dong \cite{Dong2019IROS})} & 85\% (3.04) & - & 70\% (3.34) & - & 94\% (2.60) & - & 15\% (3.83)\\
\toprule
\end{tabular}
\end{center}
\vspace{-15pt}
\label{table:result2}
\end{table*}

\subsection{RL Policy Training}

RL Policy Training is done in simulation. In every episode, to make the policy generalizable to various shapes, we randomize the object polygon by randomly scattering points and drawing their convex hull (blue polygon in Fig.\ref{fig:scatter}). An offset polygon with 2.25mm clearance is used as the hole polygon (black polygon in Fig.\ref{fig:scatter}). The initial misalignment is sampled from $\pm (12\text{mm},12\text{mm},15^{\circ}$). The simulator takes the object-hole polygon and the misalignment as input. Small Gaussian noise (0.2mm, $0.4^{\circ}$) is added to the misalignment to make the policy more robust. The simulator then computes a valid contact line by connecting intersecting points, if one exists, then a Gaussian noise (4mm, $4^{\circ}$) is added to it and returned as an output. If the object polygon intersects with the hole polygon but there is no valid contact line, the simulator returns no line. Lastly, if the object polygon lies inside the hole polygon without intersecting, the simulator returns an insertion success signal.

As a training algorithm, we use twin delayed deep deterministic policy gradient (TD3 \cite{Fujimoto}) with a recurrent actor. We formulate the task as a partially observed Markov decision process (POMDP). The observation is the estimated contact line and on which side the object tilts around the contact line. The action is the SE(2) robot displacement from the initial pose. We use the same reward function as in \cite{Dong2021}:
\vspace{-1.4pt}
\begin{equation}
    R_t = e_{t-1} - e_{t} - P + \chi R_s,
\end{equation}\vspace{-1.43pt}
where $(e_{t-1}-e_t)$ is a decrease in misalignment from the previous step, $P$ is a small constant penalty term, $R_s$ is a success reward, and $\chi$ is the success signal (1 if inserted, 0 otherwise). The maximum sequence length for each episode is set to 15. We train the policy for 15,000 episodes, which takes approximately 15 minutes on GeForce RTX 2080, while it is equivalent to about 1,000 hours of uninterrupted robot experiments.

We use 3 layers of long short-term memory (LSTM, \cite{Hochreiter}) with 64 nodes for the actor network to deal with partial observability. The LSTM alleviates the partial observability by considering the observations and actions of the current and previous steps when computing the following action. We use a multi-layer perceptron (MLP) with 64, 64, and 32 nodes in each layer for the critic. The critic takes the ground truth misalignment and the subsequent action as input and outputs the Q value estimate.

%When a valid contact line does not exists, the simulator returns `fail' with $95\%$ probability and otherwise returns a random line as a false positive. When there exists a valid contact line, the simulator returns `fail' as a false negative with some probability. This probability depends on the horizontal location of the gripper center relative to the contact line. If the gripper center lies on the opposite side of the tilting side and is far away from the contact line, it is relatively difficult to find the contact line with the push-down and the rocking motion. Therefore, we implement a sigmoid-like function as the probability of the false negative, where the probability converges to zero when the gripper center is on the same side as the tilting side and is far away from the contact line. When the simulator did not return the false negative, it returns the computed contact line with a small Gaussian noise (4mm, $4^{\circ}$) with $95\%$ probability and otherwise with a large Gaussian noise (20mm, $60^{\circ}$) to model a case where the contact sensing module makes a mistake.

\subsection{Insertion Experiment}

We test the entire framework (Contact Sensing + RL Insertion) with the real test objects and holes. For every episode, we vary grasping height and force. Initial misalignment is sampled from $\pm (12\text{mm},12\text{mm},15^{\circ}$). We evaluate the success rate and the average attempt number. The performance is compared with the same ablation models as in Section \ref{subsection:wall}.

\section{Results}

\subsection{Active Extrinsic Contact Sensing Experiment}\label{subsection:result1}

Table \ref{table:result1} shows the performance of the contact sensing experiment on the wall environment in 100 trials per object. Accuracy is the rate at which the method estimates the contact line with horizontal translation error less than 7mm and rotational error less than $25^{\circ}$. We show accuracy separately for easy cases and difficult cases. The easy cases are when the gripper center seen from the top view lies on the same side of the tilting side and the horizontal distance between the gripper center and the contact line is larger than 5mm.

Our method showed an accuracy of 95$\%$ for easy cases and 76$\%$ for difficult cases. As explained in Section \ref{subsection:iSAM}, a model without the sticking factor showed reasonable performance in the horizontal direction but the vertical error was about 3.4 times larger than the original version.

`w/o Deformation Factor' shows similar performance as the original. This is because the deformation is kept almost constant during the pivoting motion so the deformation factor becomes less important. However, `w/o Deformation Factor and w/o Pivoting Motion' showed poorer performance than `w/o Pivoting Motion' because the deformation fluctuates significantly during the push-down and the rocking motion.

For `w/o Active Motion', the accuracy drops, especially for the difficult cases. In difficult cases, the gripper center seen from the top view lies on the wall so it is difficult to tilt with only the push-down motion. In such a case, the active rocking motion is necessary to get a significant tilting angle and estimate a contact line. `w/o Control' shows the poorest performance meaning that the GelSlim compliance solely does not provide sufficient information to estimate a contact line.

\subsection{Insertion Experiments}

Table \ref{table:result2} shows the insertion performance of the proposed method, ablation models, and previous work \cite{Dong2019IROS, Dong2021}, in 100 episodes per case. The performance is evaluated in success rate and the average number of attempts until the insertion succeeds. Note that \cite{Dong2019IROS, Dong2021} used the shown cases as training cases, while ours used them as testing cases. Also, they used smaller misalignment (6mm, 6mm, $10^{\circ}$) than ours (12mm, 12mm, $15^{\circ}$).

Our method showed a higher than 95$\%$ success rate in all the test cases. Especially for the rectangle object-hole, where it requires an accurate rotation error correction, the success rate was 7$\%$ higher, and the attempt number was 24$\%$ lower than the previous work. This is because the proposed method takes informative measurement through the controlled motion and efficiently represents it as the explicit contact line estimates. It also has a recurrent structure, which is influenced by estimates of previous steps.

`w/o Sticking Factor' shows similar performance as the original since it has a similar contact line estimation performance in the horizontal direction as seen in Section \ref{subsection:result1}. The performance decreases slightly for `w/o Deformation Factor'. `w/o Active Motion' shows poorer performance, especially for the rectangle object-hole case. %`w/o Control' showed the poorest performance.

\section{Conclusion and Future Work}

We propose a framework for active extrinsic contact sensing and apply it to an insertion policy by using the estimated contact line as an input to an insertion policy trained in simulation with RL. The factor graph-based estimation model and the active tactile feedback controller work collaboratively to localize the contact line between a grasped object and an environment. We then formulate the insertion task as an RL problem where the input is the series of estimated contacts with the hole. This enables us to train the RL agent in simulation and ease the burden of collecting training data with real experiments. In future work, we would like to extend this framework to more general manipulation scenarios: point contacts, non-stationary contacts, and non-flat object/environment surfaces.

%\addtolength{\textheight}{-12cm}

\bibliography{IEEEabrv,ICRASK}
\bibliographystyle{IEEEtran}

\end{document}